\documentclass[10pt,conference]{IEEEtran}

\usepackage{graphicx}
\usepackage{amsmath}
\usepackage{amssymb}

\begin{document}

\title{Evaluating Hybrid Quantum-Classical Models for Reduced-Order Brain Deformation Dynamics}


\author{
\IEEEauthorblockN{
Tao Liu\IEEEauthorrefmark{1},
Ge He\IEEEauthorrefmark{1},
Dongyu Liang\IEEEauthorrefmark{1},
and Wujie Wen\IEEEauthorrefmark{2}
}
\IEEEauthorblockA{
\IEEEauthorrefmark{1}\textit{Lawrence Technological University},
\IEEEauthorrefmark{2}\textit{North Carolina State University}\\
\IEEEauthorrefmark{1}\texttt{\{tliu3, ghe, dliang\}@ltu.edu},
\IEEEauthorrefmark{2}\texttt{wwen2@ncsu.edu}
}
\vspace{-2em}
}

\maketitle

\begin{abstract}
We evaluate hybrid quantum-classical machine learning for the reduced-order prediction of spatiotemporal brain deformation fields. To mitigate the computational intractability of high-dimensional displacement fields, we employ Proper Orthogonal Decomposition (POD) to project the data into a compact latent space. Within this framework, we formulate two distinct learning objectives: static temporal-to-latent regression and autoregressive latent state forecasting. We systematically benchmark compact classical baselines against both minimal and enhanced hybrid quantum architectures. 
Our results demonstrate that classical networks provide the strongest baselines in the present setting. For static regression, a classical POD-MLP outperforms all evaluated quantum variants, although an enhanced Variational Quantum Circuit (VQC) substantially improves upon a minimal VQC baseline. For temporal forecasting, a classical POD-LSTM delivers superior predictive accuracy and statistical robustness compared to an enhanced Quantum LSTM (QLSTM) across varying history windows and random initializations. Overall, this study establishes reduced-order physical field learning as a rigorous testbed for near-term QML, highlighting that while hybrid enhancements successfully recover expressivity in weak quantum circuits, classical architectures retain a definitive advantage in both fidelity and stability.
\end{abstract}


\section{Introduction}
\label{sec:intro}

Hybrid quantum-classical machine learning (QML) is a promising tool for scientific computing, yet a significant practical gap remains. Current variational quantum models are largely constrained to low-dimensional inputs, whereas real-world biomechanics and engineering data are inherently high-dimensional, spatially distributed, and time-dependent. In classical computing, this dimensionality bottleneck is typically resolved using reduced-order modeling (ROM), such as proper orthogonal decomposition (POD). While data-driven ROMs are standard in biomechanics~\cite{wu2024mlbiomech,han2024organdeformation,farenga2025latent,li2025latentpde}, their intersection with QML remains largely unexplored for realistic spatiotemporal field prediction~\cite{peralgarcia2024qmlreview,benedetti2019pqc,mitarai2018qcl,perezsalinas2020reuploading,mohammadi2025qmlsurrogate}.

To bridge this gap, we formulate a QML-compatible framework using brain deformation prediction as a complex physical testbed. By projecting the full spatiotemporal displacement field into a compact POD latent space, we convert the massive full-field problem into two tractable learning tasks: a static regression task (mapping individual timesteps to latent states) and a temporal forecasting task (predicting future states from historical windows). This formulation makes the problem computationally accessible for quantum models while preserving predictive evaluation in the reconstructed physical space.

Rather than proposing a single model, we establish a structured comparative framework spanning compact classical baselines, minimal quantum baselines, and enhanced hybrid quantum variants. This allows us to investigate two fundamental questions: First, in a smooth reduced-order regression task, does quantum enhancement offer tangible advantages over strong classical models? Second, in an autoregressive setting, can a hybrid quantum bottleneck effectively capture latent physical dynamics, and does deeper hybridization improve trainability and robustness?

The main contributions of this work are as follows:
\begin{itemize}
\item We study brain deformation prediction in a QML-compatible reduced-order setting by projecting the full displacement field onto a compact POD latent space.
\item We establish a structured comparison across compact classical baselines, minimal quantum baselines, and enhanced hybrid quantum models for both static regression and temporal forecasting.
\item We show experimentally that hybrid enhancements can substantially improve weak quantum baselines, while compact classical models remain the strongest and most robust performers in the present setting.
\item We provide practical guidance for applying QML to reduced-order scientific field prediction, with emphasis on physical-space reconstruction, robustness, and sensitivity to temporal context.
\end{itemize}
\section{Methodology: Reduced-Order Formulation and Hybrid Quantum Architectures}
\label{sec:meth}

We consider a brain deformation prediction problem defined on a decimated spatiotemporal mesh. At each timestep, the brain is represented by a set of mesh nodes with spatial coordinates $(x,y,z)$ and displacement components $(u_x,u_y,u_z)$. Thus, the deformation state of the mesh can be viewed as a time-evolving vector field defined over the brain geometry. Our goal is to learn this deformation behavior from a single loading trajectory and to predict either the reduced deformation state itself or its temporal evolution.

\begin{figure}[t]
    \centering
    \includegraphics[width=\columnwidth, trim={0 0 0 5em}, clip]{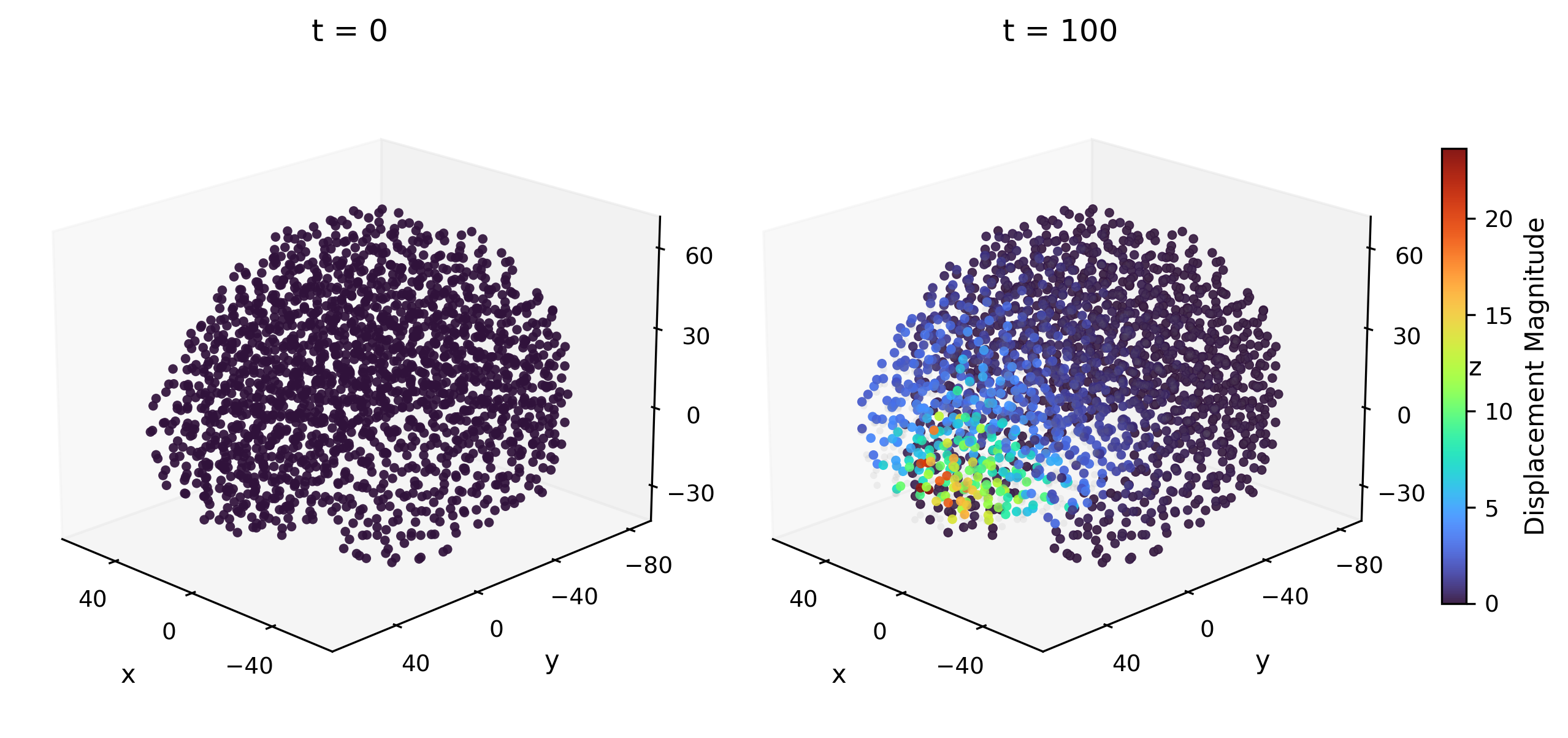}
    \vspace{-2em}
    \caption{Brain deformation field on the decimated mesh at the initial (left) and final (right) timesteps. Node color indicates displacement magnitude (mm).}
    \label{fig:brain_mesh_deformation}
    \vspace{-1em}
\end{figure}

Fig.~\ref{fig:brain_mesh_deformation} illustrates the deformation field on the decimated brain mesh at the initial and final timesteps, highlighting that the prediction target is a full spatial displacement field evolving over time rather than a scalar response. Direct learning in the original field space is therefore high-dimensional and computationally demanding, especially for current near-term hybrid quantum-classical models~\cite{bharti2022noisy}.

To make the problem tractable, we introduce a reduced-order latent representation of the deformation field using Proper Orthogonal Decomposition (POD)~\cite{berkooz1993proper,brunton2019data}, allowing us to perform learning in a compact space. Let $\mathbf{z}_t \in \mathbb{R}^{k}$ denote the POD latent coefficients of the displacement field at timestep $t$. After projection-based reduction, the original high-dimensional brain deformation problem is reformulated as learning in a compact latent space. Within this representation, we study two related but distinct prediction tasks:
$$t \mapsto \hat{\mathbf{z}}_t$$
for static reduced-order regression, and
$$(\mathbf{z}_{t-w+1}, \ldots, \mathbf{z}_t) \mapsto \hat{\mathbf{z}}_{t+1}$$
for temporal reduced-order forecasting, where $w$ is the history window size.

The selected model families are not intended to exhaust all possible architectures. Instead, they define a structured comparison ladder under a common reduced-order formulation, including compact classical baselines, minimal quantum baselines, and stronger hybrid quantum variants. This design separates reduced-order representation learning from temporal transition learning, addressing two core questions: First, in the static setting, does quantum enhancement provide value for direct reduced-order regression, or does it primarily compensate for weak minimal quantum formulations? Second, in the temporal setting, can a quantum bottleneck support latent dynamics forecasting, and does stronger hybridization improve trainability and forecasting quality?

\subsection{Static Model Families}

The static model family is introduced as a lower-complexity control setting. By removing recurrent memory and autoregressive rollout, it isolates the ability of each model to approximate the reduced deformation manifold itself. If a model struggles in the static setting, weak temporal performance may reflect limited representational power rather than a failure to learn dynamics. We consider four models within this setting:

\subsubsection{Standard POD-MLP}
A compact classical multilayer perceptron that maps the normalized timestep directly to the latent coefficients. It consists of two fully connected hidden layers with \texttt{tanh} activations, followed by a linear output layer producing the $k$-dimensional latent vector. This serves as the primary classical baseline for smooth reduced-order regression.

\subsubsection{Quantum-Enhanced POD-MLP}
This model augments the feedforward pipeline with a hybrid quantum feature transformation. The scalar timestep is lifted to an $n_q$-dimensional bounded feature vector through a classical input layer. These features are encoded into a parameterized quantum circuit using angle embedding and processed by trainable entangling layers~\cite{benedetti2019pqc}. The resulting qubit expectation values are mapped back to the latent space through a classical output layer. This tests whether a quantum feature transformation adds value even when the underlying regression task is well-approximated classically.

\subsubsection{Standard POD-VQC}
A minimal memoryless variational quantum regressor (VQC)~\cite{cerezo2021variational}. The scalar timestep is broadcast to all qubits, encoded once, processed by a shallow variational circuit, and measured directly to produce the latent prediction. This serves as our absolute minimal quantum baseline.

\subsubsection{Enhanced POD-VQC}
This extends the minimal VQC design with three elements: a classical input dressing layer, repeated embedding of the input across circuit depth (leveraging the data re-uploading paradigm~\cite{perezsalinas2020reuploading}), and a classical output dressing layer. This comparison tests whether hybrid enhancement can recover the expressivity lost in a shallow minimal quantum formulation.

\subsection{Temporal Model Families}

The temporal model family addresses the prediction of latent dynamics rather than direct fitting of the latent manifold. The model must infer the update law on the reduced-order state, with forecasting quality affected by autoregressive rollout and error accumulation. We compare one classical recurrent baseline and two hybrid recurrent variants:

\subsubsection{Classical POD-LSTM}
Uses a standard Long Short-Term Memory (LSTM) encoder to process the latent history window $(\mathbf{z}_{t-w+1}, \ldots, \mathbf{z}_t)$. The final hidden state is mapped to the next latent vector through a classical output head. This is the reference recurrent surrogate for latent dynamics forecasting.

\subsubsection{Standard POD-QLSTM}
Adapting the Quantum LSTM framework~\cite{chen2020qlstm}, this replaces the classical output head with a shallow quantum bottleneck. After latent sequence encoding by the LSTM, the final hidden state is projected to qubit dimension, bounded by a \texttt{tanh} activation, and encoded into a VQC. The measured expectation values are used directly as the next latent prediction, testing if a shallow quantum bottleneck can support reduced-order temporal prediction.

\subsubsection{Enhanced POD-QLSTM}
Augments the recurrent hybrid design with a stronger quantum-classical interface. The final LSTM hidden state passes through a classical pre-quantum bridge to produce bounded features, which are repeatedly embedded into a deeper VQC. Measurements are then mapped to the latent space via a classical output dressing network. This tests whether deeper hybridization mitigates the representational weakness and recurrent optimization difficulties of the minimal QLSTM.

In all cases, latent predictions are mapped back to the physical displacement field through inverse POD reconstruction and inverse scaling, ensuring model behavior is assessed both in latent space and reconstructed physical space.
\section{Experimental Setup}
\label{sec:exp}

\subsection{Dataset Preparation}

We utilized a spatiotemporal dataset representing a decimated brain mesh subjected to a single loading trajectory at the front of the brain. The mesh comprises $N = 1940$ nodes. At each of the $T = 51$ timesteps, the state is defined by the nodes' spatial coordinates $(x,y,z)$ and their corresponding displacement vectors $(u_x,u_y,u_z)$. To format this data for model ingestion, the displacement field at each timestep was flattened into a single vector by concatenating all node-wise components in the order $(u_x,u_y,u_z)$, yielding a full-field state matrix $\mathbf{X} \in \mathbb{R}^{T \times 3N}$. 

We employed a strict temporal extrapolation split: the first 46 snapshots were used for training, while the final 5 were held out for testing. This split is intentionally rigorous; it evaluates the models' ability to forecast unseen future deformation states rather than merely interpolating between previously observed ones.

\subsection{Reduced-Order Preprocessing}

Given the high dimensionality of the full-field data, all models were trained and evaluated in a reduced latent space. To prevent the leakage of future information into the preprocessing pipeline, displacement features were first normalized to the range $[-1, 1]$ using a \texttt{MinMaxScaler} fitted exclusively on the training snapshots. The identical scaling transformation was then applied to the held-out test set. 

Following normalization, we constructed a reduced-order latent representation using POD~\cite{berkooz1993proper, brunton2019data}, fitted solely on the scaled training matrix. Projecting both training and test snapshots onto the leading modes yields a compact representation $\mathbf{z}_t \in \mathbb{R}^{k}$. Preliminary variance analysis confirmed that the deformation sequence is strongly low-rank, with the leading modes capturing the vast majority of the cumulative variance. 

For all subsequent classical and hybrid quantum experiments, we fixed the latent dimension to $k = 8$. This dimension was selected not as the intrinsic optimal rank, but as a practical benchmark size that remains small enough to accommodate near-term QML constraints while retaining the dominant structural variance of the deformation trajectory. Ultimately, all model predictions were mapped back to the physical space via inverse POD reconstruction and inverse scaling to assess real-world fidelity.

\subsection{Training and Evaluation Protocol}

We evaluated the models across the two tasks defined in Section~\ref{sec:meth}.
For the temporal task, test-set predictions were generated via autoregressive rollout, where each predicted state was iteratively fed back into the input window.

All models were trained for 500 epochs using the Adam optimizer with a learning rate of $0.01$. The recurrent architectures utilized a single LSTM layer with a hidden dimension of 16. In the static experiments, all models were trained under identical temporal splits and latent representations, ensuring that performance discrepancies isolated architectural behaviors rather than preprocessing artifacts.

To rigorously assess stability in the temporal setting—given that recurrent hybrid quantum models can be highly sensitive to initialization and context length—we conducted a robustness sweep. We evaluated history windows $w \in \{3, 5, 10\}$ and repeated each experiment across five random seeds, aggregating the results as mean $\pm$ standard deviation. 

Model performance was quantified using the coefficient of determination ($R^2$) evaluated both in the latent space (on predicted POD coefficients) and in the reconstructed physical space (on displacement components $U_x$, $U_y$, and $U_z$). Furthermore, we tracked convergence behavior through loss trajectories and monitored gradient flow diagnostics for the QLSTM variants to analyze optimization dynamics within the recurrent encoder.

All experiments were executed on a workstation equipped with an AMD Ryzen Threadripper 3970X CPU, 256\,GB of RAM, and an NVIDIA A100 GPU (80\,GB VRAM). The classical neural networks and overall training pipelines were implemented in PyTorch, while the quantum machine learning components were constructed and simulated using PennyLane.

\section{Results and Discussion}
\label{sec:eval}

\begin{table}[t]
\centering
\caption{Static reduced-order regression results.}
\label{tab:static_results}
\vspace{-0.5em}
\begin{tabular}{lcccc}
\hline
Model & $Z$ $R^2$ & $U_x$ $R^2$ & $U_y$ $R^2$ & $U_z$ $R^2$ \\
\hline
Standard POD-MLP            & 0.7812 & 0.9728 & 0.9794 & 0.9452 \\
Quantum-Enhanced POD-MLP    & 0.7151 & 0.9304 & 0.9316 & 0.8739 \\
Standard POD-VQC            & 0.2003 & 0.7803 & 0.8065 & 0.6326 \\
Enhanced POD-VQC            & 0.4904 & 0.9186 & 0.9267 & 0.8790 \\
\hline
\end{tabular}
\end{table}

\begin{figure}[t]
    \centering
    \includegraphics[width=\columnwidth]{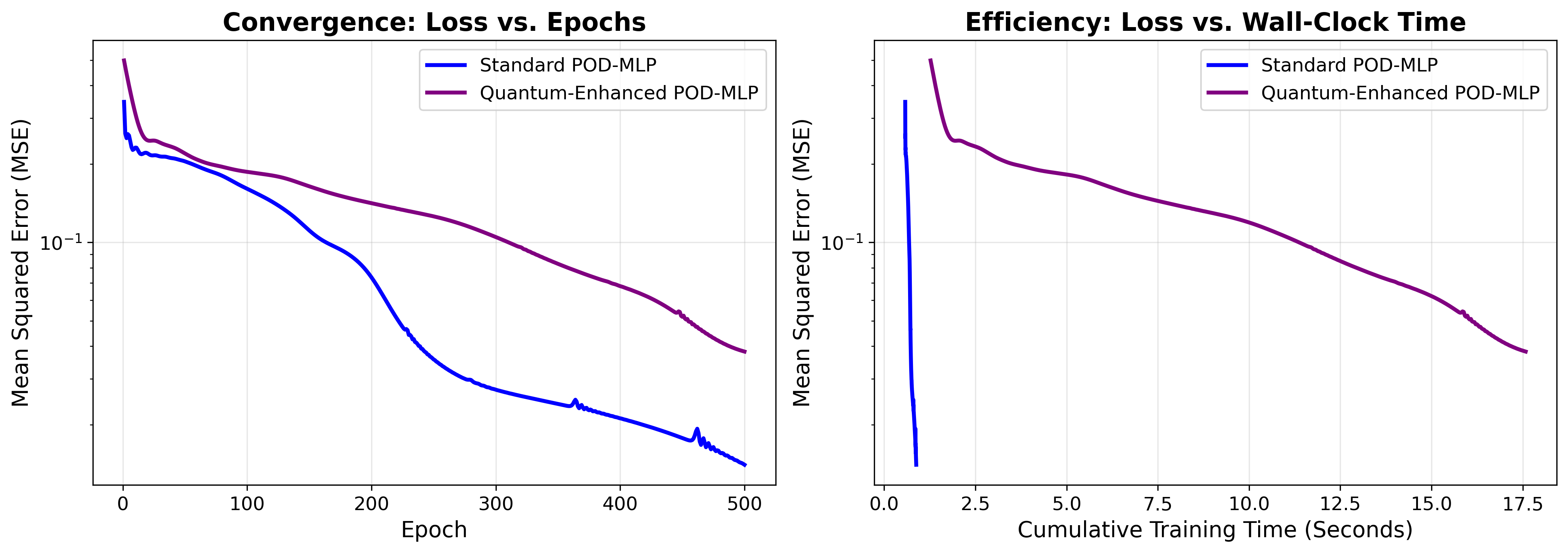}
    \includegraphics[width=\columnwidth]{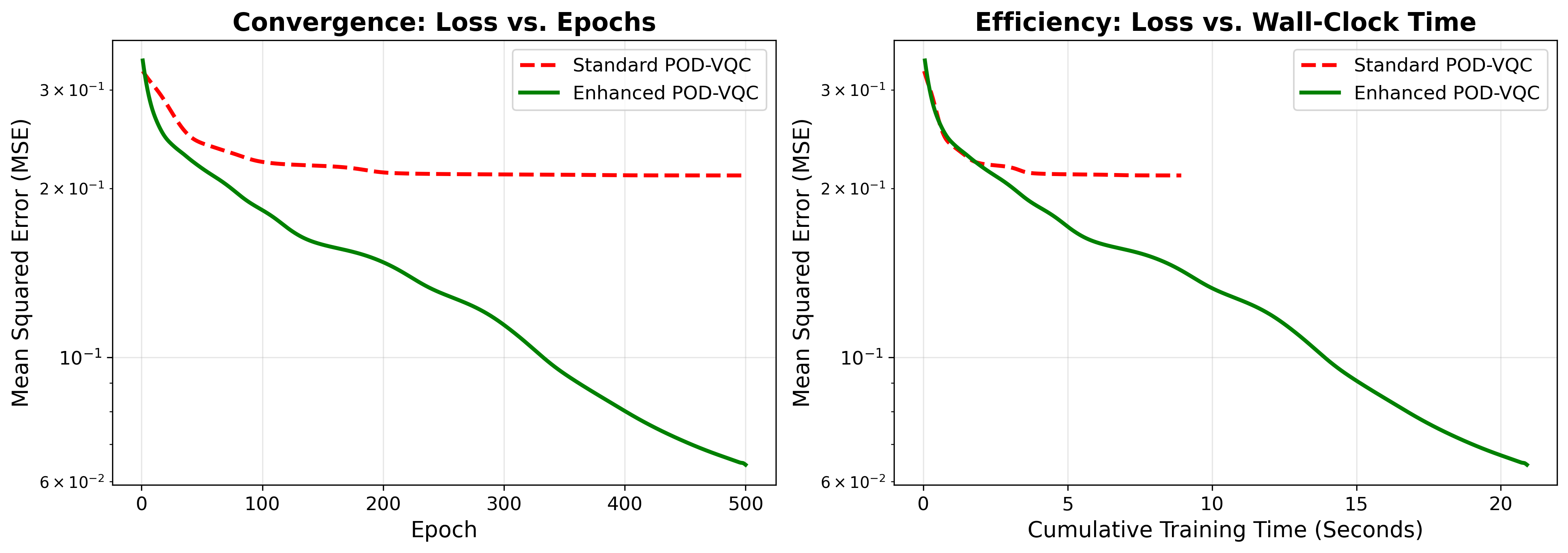}
    \vspace{-2em}
    \caption{Static-family convergence and efficiency comparisons. Top: Standard POD-MLP versus Quantum-Enhanced POD-MLP. Bottom: Standard POD-VQC versus Enhanced POD-VQC.}
    \label{fig:static_convergence}
    \vspace{-1em}
\end{figure}

\begin{figure}[t]
    \centering
    \includegraphics[width=\columnwidth, trim={0 0 0 5em}, clip]{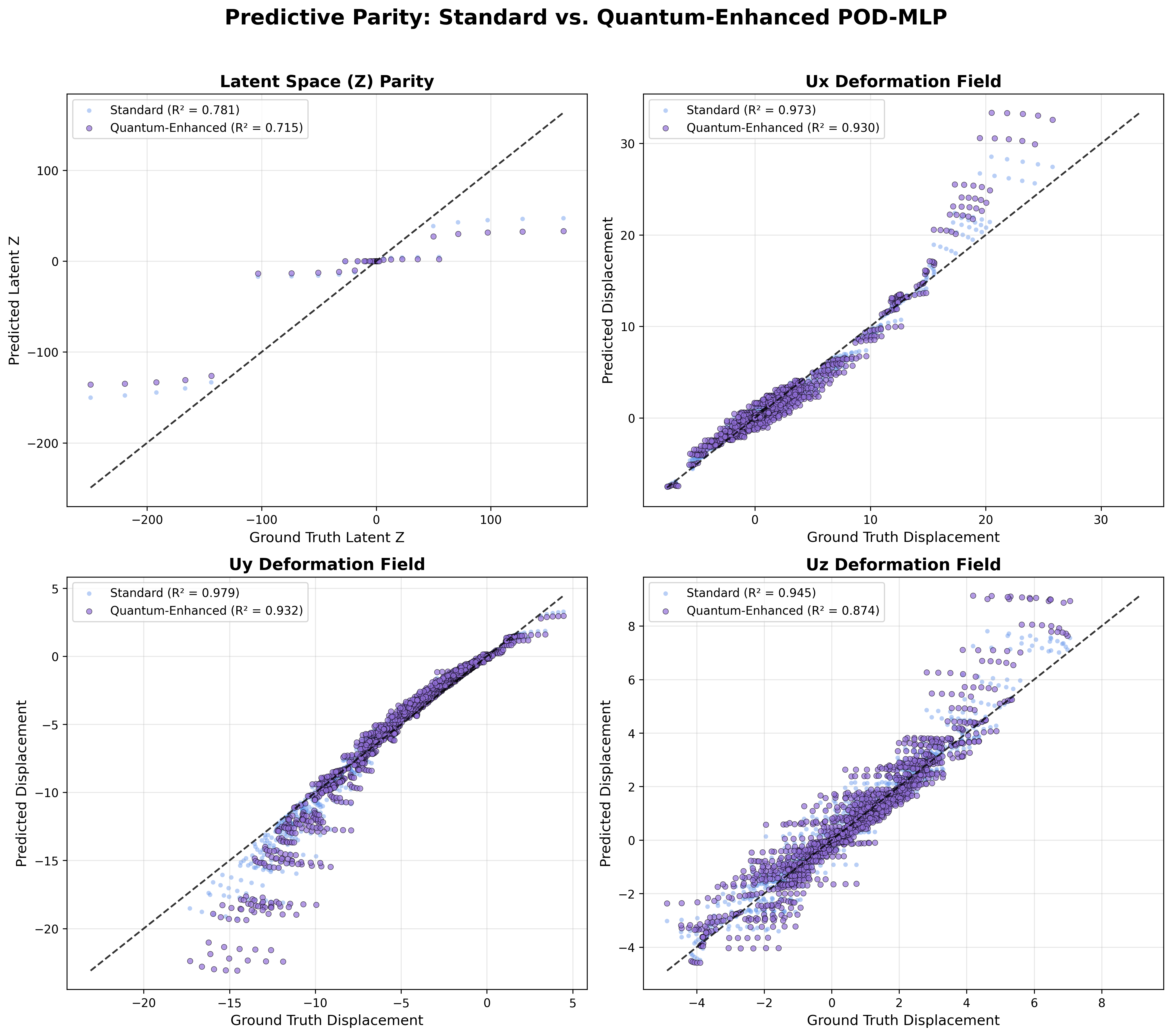}
    \vspace{-2em}
    \caption{Predictive parity comparison for the POD-MLP family.}
    \label{fig:pod_mlp_parity}
    \vspace{-1em}
\end{figure}

\begin{figure}[t]
    \centering
    \includegraphics[width=\columnwidth, trim={0 0 0 5em}, clip]{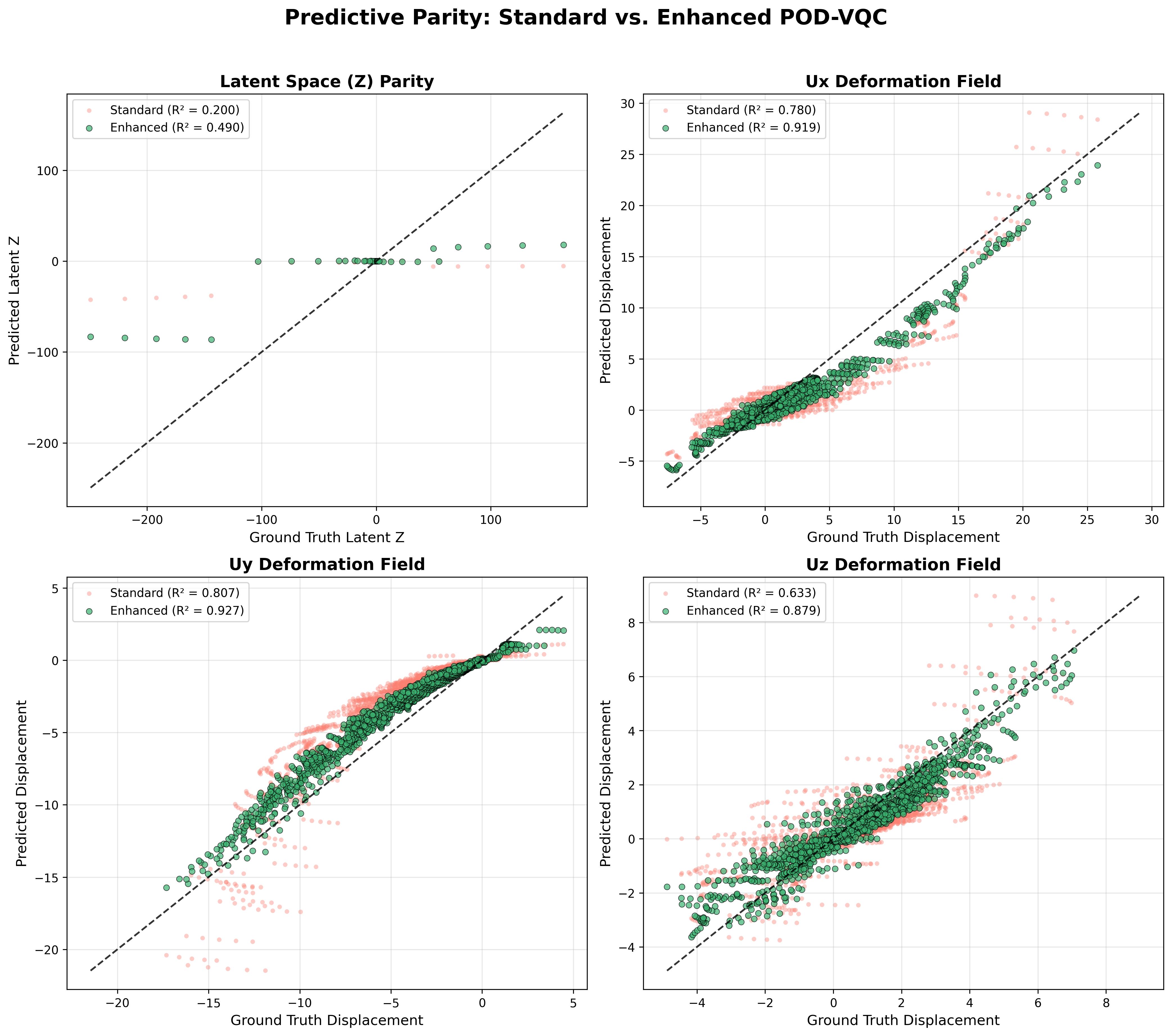}
    \vspace{-2em}
    \caption{Predictive parity plots for the POD-VQC family.}
    \label{fig:pod_vqc_parity}
    \vspace{-1em}
\end{figure}

\subsection{Static Reduced-Order Regression}

We first evaluate the static reduced-order regression task, wherein each model predicts the latent state directly from the normalized timestep. Quantitative metrics are summarized in TABLE~\ref{tab:static_results}, while predictive parity plots for the MLP and VQC architectures are presented in Fig.~\ref{fig:pod_mlp_parity} and Fig.~\ref{fig:pod_vqc_parity}, respectively.

Among all evaluated architectures, the Standard POD-MLP achieved the highest overall performance. It yielded a latent-space coefficient of determination ($R^2$) of 0.7812, alongside highly accurate physical reconstructions with $R^2$ values of 0.9728, 0.9794, and 0.9452 for the $U_x$, $U_y$, and $U_z$ displacement fields, respectively. Although the Quantum-Enhanced POD-MLP successfully learned the temporal-to-latent mapping, it consistently underperformed its classical counterpart across all metrics (latent $R^2 = 0.7151$; physical $R^2 = 0.9304$, $0.9316$, and $0.8739$). As illustrated in Fig.~\ref{fig:pod_mlp_parity}, the classical MLP demonstrates noticeably tighter alignment with the ground truth. This suggests that for smooth, single-trajectory regression in a low-rank space, a compact classical feedforward network possesses sufficient expressivity, rendering quantum feature transformations largely unnecessary for this specific task.

A contrasting dynamic emerges within the VQC family. The Standard POD-VQC, acting as a minimal memoryless quantum baseline, exhibited poor predictive capability, achieving a latent $R^2$ of only 0.2003 alongside correspondingly weak physical reconstructions. However, the Enhanced POD-VQC yielded substantial improvements, elevating the latent $R^2$ to 0.4904 and the physical reconstruction scores to 0.9186, 0.9267, and 0.8790 for $U_x$, $U_y$, and $U_z$, respectively. This drastic improvement is visually evident in Fig.~\ref{fig:pod_vqc_parity}. Thus, while hybrid enhancement proves highly effective at recovering expressivity lost in minimal quantum circuits, the enhanced VQC still falls short of the Standard POD-MLP baseline.

A notable phenomenon across all models is that physical-space reconstruction quality is systematically higher than the latent-space fit would suggest. This disparity highlights that moderate errors in the predicted latent coordinates do not necessarily translate into severe physical distortion; because the leading POD modes capture the dominant macroscopic deformation patterns, the inverse mapping is somewhat forgiving to minor latent inaccuracies. Furthermore, predicting the $U_z$ component consistently proved the most challenging across all architectures, indicating that deformation in the third dimension exhibits more complex, harder-to-compress dynamics than the $U_x$ and $U_y$ movements.

Ultimately, these static results emphasize that architectural suitability is more critical than the mere inclusion of quantum parameters. In this reduced-order regression setting, compact classical models remain the most robust and practical choice. Hybrid quantum techniques, rather than demonstrating a quantum advantage, primarily serve to rescue otherwise underpowered, shallow quantum baselines.

\begin{figure}[t]
    \centering
    \includegraphics[width=\columnwidth]{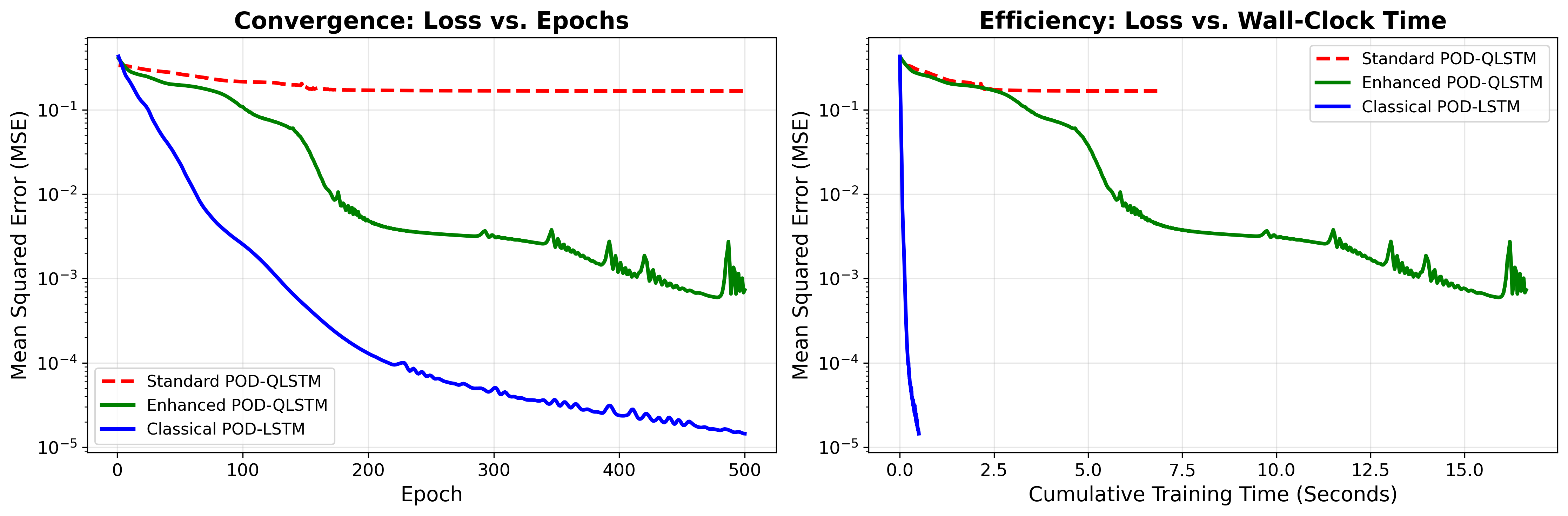}
    \vspace{-2em}
    \caption{Convergence and efficiency comparison for the recurrent model family.}
    \label{fig:recurrent_convergence}
    \vspace{-1em}
\end{figure}

\begin{figure}[t]
    \centering
    \includegraphics[width=\columnwidth, trim={0 0 0 5em}, clip]{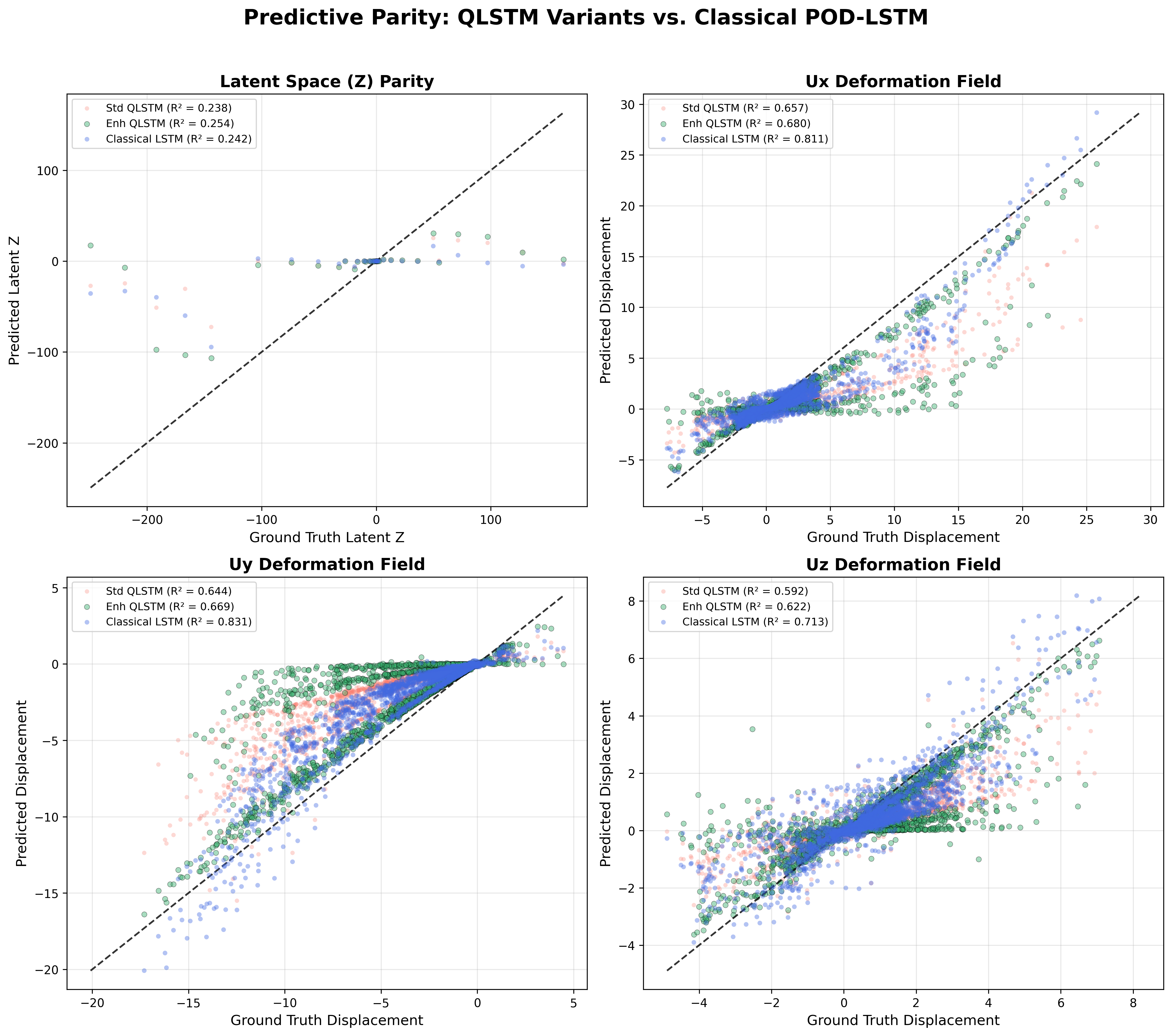}
    \vspace{-2em}
    \caption{Predictive parity plots for the recurrent model family.}
    \label{fig:recurrent_parity}
    \vspace{-1em}
\end{figure}

\begin{figure}[t]
    \centering
    \includegraphics[width=\columnwidth, trim={0 0 0 0}, clip]{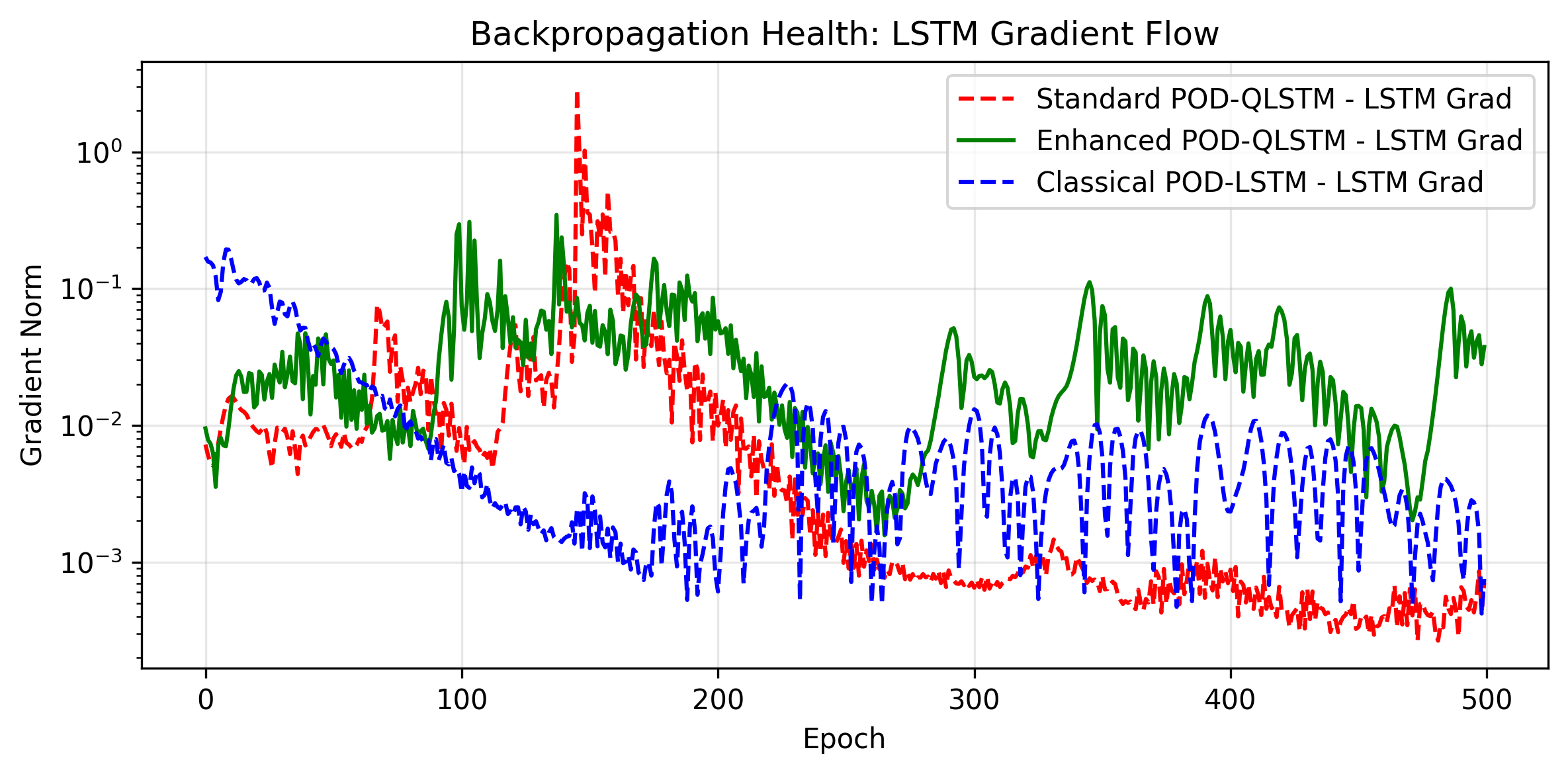}
    \vspace{-2em}
    \caption{Gradient-flow diagnostics for the recurrent hybrid models.}
    \label{fig:recurrent_gradient_flow}
    \vspace{-2em}
\end{figure}

\begin{table*}[t]
\centering
\caption{Window-size \(\times\) seed sweep. Results are reported as mean \(\pm\) standard deviation \textbf{over 5 random seeds}.}
\label{tab:recurrent_sweep}
\vspace{-1em}
\small
\setlength{\tabcolsep}{6pt}
\begin{tabular}{c l c c c c c}
\hline
Window & Model & $Z$ $R^2$ & $U_x$ $R^2$ & $U_y$ $R^2$ & $U_z$ $R^2$ \\
\hline
3  & Classical POD-LSTM & $0.074 \pm 0.101$ & $0.584 \pm 0.150$ & $0.570 \pm 0.167$ & $0.506 \pm 0.155$ \\
3  & Standard POD-QLSTM & $0.133 \pm 0.086$ & $0.565 \pm 0.229$ & $0.540 \pm 0.283$ & $0.441 \pm 0.165$ \\
3  & Enhanced POD-QLSTM & $-0.016 \pm 0.218$ & $0.436 \pm 0.406$ & $0.413 \pm 0.466$ & $0.313 \pm 0.390$ \\
\hline
5  & Classical POD-LSTM & $0.163 \pm 0.097$ & $0.683 \pm 0.139$ & $0.681 \pm 0.154$ & $0.606 \pm 0.143$ \\
5  & Standard POD-QLSTM & $0.109 \pm 0.084$ & $0.527 \pm 0.206$ & $0.490 \pm 0.268$ & $0.417 \pm 0.142$ \\
5  & Enhanced POD-QLSTM & $0.192 \pm 0.208$ & $0.685 \pm 0.211$ & $0.685 \pm 0.242$ & $0.587 \pm 0.199$ \\
\hline
10 & Classical POD-LSTM & $0.303 \pm 0.087$ & $0.814 \pm 0.055$ & $0.826 \pm 0.055$ & $0.738 \pm 0.070$ \\
10 & Standard POD-QLSTM & $0.124 \pm 0.198$ & $0.554 \pm 0.145$ & $0.525 \pm 0.165$ & $0.447 \pm 0.128$ \\
10 & Enhanced POD-QLSTM & $0.137 \pm 0.288$ & $0.555 \pm 0.294$ & $0.534 \pm 0.340$ & $0.450 \pm 0.270$ \\
\hline
\end{tabular}
\vspace{-2em}
\end{table*}





\subsection{Temporal Reduced-Order Forecasting}

We next evaluate the temporal reduced-order forecasting task, where models predict the subsequent latent state from a historical window and generate future trajectories via autoregressive rollout. Fig.~\ref{fig:recurrent_convergence} illustrates the convergence and computational efficiency of the recurrent models, while predictive parity is shown in Fig.~\ref{fig:recurrent_parity}. Additionally, Fig.~\ref{fig:recurrent_gradient_flow} visualizes the gradient-flow dynamics within the QLSTM variants, and TABLE~\ref{tab:recurrent_sweep} aggregates the robustness sweep across varying window sizes and random seeds.

In a representative single-run comparison, the Classical POD-LSTM delivered the strongest predictive performance. It achieved a latent-space $R^2$ of 0.2418 and physical-space reconstruction $R^2$ values of 0.8110, 0.8305, and 0.7133 for $U_x$, $U_y$, and $U_z$, respectively. Although the Standard and Enhanced POD-QLSTM architectures achieved comparable latent-space accuracy, both fell short of the classical baseline in the reconstructed physical space. As depicted in Fig.~\ref{fig:recurrent_parity}, the Classical POD-LSTM maintains the tightest alignment with the ground truth. While the Enhanced POD-QLSTM consistently outperforms its minimal standard counterpart—confirming the value of hybrid enhancement—it remains insufficient to surpass the classical recurrent baseline.

Convergence trajectories and efficiency metrics corroborate these findings. As shown in Fig.~\ref{fig:recurrent_convergence}, the Classical POD-LSTM converges more rapidly and descends to a lower final training loss than either hybrid alternative. The Standard POD-QLSTM plateaus early, firmly establishing itself as the weakest recurrent model. Conversely, the Enhanced POD-QLSTM demonstrates sustained improvement throughout training, narrowing the performance gap with the classical baseline, albeit without overtaking it. This indicates that deeper hybrid integration successfully alleviates optimization bottlenecks present in minimal QLSTMs, but does not fundamentally alter the overall performance hierarchy.

Gradient-flow diagnostics (Fig.~\ref{fig:recurrent_gradient_flow}) provide further insight into this optimization behavior. The Enhanced POD-QLSTM maintains stronger and more sustained gradient signals through its recurrent encoder compared to the minimal Standard POD-QLSTM. This validates the intended architectural function of the classical pre-quantum bridge and output dressing layers. However, this improved gradient propagation does not strictly translate into superior final predictive accuracy. This disconnect suggests that the performance ceiling of recurrent hybrid models in this setting is dictated not solely by optimization difficulties, but also by inherent model-task mismatches and the compounding nature of autoregressive forecasting errors.

To rigorously assess stability beyond a single run, we conducted a robustness sweep over history windows $w \in \{3,5,10\}$ across five random initializations per model. As detailed in TABLE~\ref{tab:recurrent_sweep}, the Classical POD-LSTM emerged as the most robust architecture overall. Its physical-space accuracy scales reliably with longer contextual histories, peaking at $w=10$ with mean $R^2$ values of $0.814 \pm 0.055$, $0.826 \pm 0.055$, and $0.738 \pm 0.070$ for $U_x$, $U_y$, and $U_z$, respectively. In stark contrast, the Standard POD-QLSTM remains uniformly weak across all evaluated window sizes.

The Enhanced POD-QLSTM exhibits highly context-dependent behavior. At a medium window of $w=5$, it becomes surprisingly competitive with the Classical POD-LSTM, yielding comparable mean performance in $U_x$ and $U_y$. However, this parity fails to generalize: performance degrades sharply at $w=3$ and shows no further benefit at $w=10$. Furthermore, both QLSTM variants display substantially larger standard deviations across seeds compared to the classical baseline, highlighting a heightened sensitivity to initialization and a general lack of operational robustness.

Ultimately, these temporal results demonstrate that recurrent hybrid quantum models are considerably more fragile than their static counterparts when applied to reduced-order physical dynamics. While architectural enhancements successfully fortify minimal QLSTM baselines and unlock competitive performance in narrow, specific regimes, the Classical POD-LSTM remains unequivocally the strongest, most reliable, and most scalable temporal model across varying contexts and initializations.

\subsection{Lessons for Reduced-Order Spatiotemporal QML}

Synthesizing our experimental findings, we outline several actionable principles for applying hybrid QML to reduced-order spatiotemporal field learning:

\textbf{1. Projection-based reduction is a mandatory bridge for high-dimensional QML.}
The target displacement field's spatial dimension ($3N$) far exceeds the capacity of current variational quantum models. The POD latent bridge is thus a foundational necessity: it compresses the full-field prediction problem into a tractable lower-dimensional task while preserving the ability to evaluate physical fidelity. This demonstrates that near-term QML becomes viable for scientific computing only through rigorous projection into a compact latent space.

\textbf{2. Compact classical networks establish formidable baselines for smooth regression.}
As observed, the Standard POD-MLP achieved the strongest overall static performance, eclipsing both the minimal and enhanced quantum variants. This indicates that when the static mapping $t \mapsto \mathbf{z}_t$ is smoothly structured and low-dimensional, a lightweight classical feedforward model already provides an excellent approximation, setting a rigorous performance floor that is difficult to beat.

\textbf{3. Hybrid enhancement primarily rescues under-expressive quantum baselines.}
The clearest benefit of quantum hybridization emerged within the VQC and QLSTM families, where deep enhancements (e.g., data re-uploading and classical dressing) drastically improved upon minimal baselines. However, these modifications served primarily as a critical intervention to recover expressivity in otherwise underpowered quantum circuits, rather than a mechanism to achieve quantum supremacy over classical counterparts.

\textbf{4. Temporal quantum models are highly vulnerable to compounded autoregressive errors.}
The recurrent setting proved markedly more fragile than the static one. Although the Enhanced POD-QLSTM preserved stronger gradient signals during optimization, this did not strictly translate to superior predictive accuracy. The QLSTM variants exhibited significant variance and sensitivity to contextual history, highlighting the severe numerical challenge of maintaining stability against the rapid accumulation of errors during autoregressive rollout.

\textbf{5. Evaluation must prioritize physical reconstruction and statistical robustness.}
A fundamental methodological lesson emerges from the disparity between latent and physical metrics. In static tasks, moderate latent errors still yielded strong physical reconstructions via dominant POD modes; conversely, in temporal tasks, minor latent discrepancies compounded into severe physical-space degradation. Thus, future QML applications must anchor their evaluation in reconstructed physical fidelity and statistical robustness across initializations, rather than relying solely on latent-space goodness-of-fit.

Overall, our results do not indicate a clear advantage for QML in the present setting. Instead, they suggest that QML is most useful when applied within a reduced-order framework and assessed against strong classical baselines using physically meaningful metrics.

\section{Conclusion}
\label{sec:con}

We investigated hybrid quantum-classical machine learning for brain deformation prediction by employing a POD latent bridge to compress high-dimensional displacement fields into tractable static regression and temporal forecasting tasks. Our evaluations demonstrate that compact classical networks currently establish the performance ceiling: the classical POD-MLP and POD-LSTM consistently outperformed all quantum variants in static and temporal settings, respectively. Although deep hybridization successfully recovered expressivity for minimal quantum baselines, it remained insufficient to surpass classical models in this low-rank regime. Ultimately, while this reduced-order framework provides a rigorous testbed for near-term QML in biomechanics, future research must explore whether quantum architectures unlock definitive advantages in more complex scenarios, such as multi-trajectory loading or non-smooth transient dynamics.

\bibliographystyle{IEEEtran}
\bibliography{IEEEexample}

\end{document}